# Evaluating Prompt Engineering Techniques for RAG in Small Language Models: A Multi-Hop QA Approach

Amir Hossein Mohammadi
Department of Software Engineering
University of Isfahan
Isfahan Iran
a.mohammadighahderijani@eng.ui.ac.ir

Ali Moeinian
Department of Software Engineering
University of Isfahan
Isfahan, Iran
a.moeinian@eng.ui.ac.ir

Zahra Razavizade
Department of Software Engineering
University of Isfahan
Isfahan, Iran
Zahrarazavizade02@eng.ui.ac.ir

Afsaneh Fatemi
Department of Software Engineering
Faculty of Computer Engineering
University of Isfahan
Isfahan, Iran
a_fatemi@eng.ui.ac.ir

Reza Ramezani
Department of Software Engineering
Faculty of Computer Engineering
University of Isfahan
Isfahan, Iran
r.ramezani@eng.ui.ac.ir

*Abstract* — **Retrieval Augmented Generation (RAG) is a powerful approach for enhancing the factual grounding of language models by integrating external knowledge. While widely studied for large language models, the optimization of RAG for Small Language Models (SLMs) remains a critical research gap, particularly in complex, multi-hop question-answering tasks that require sophisticated reasoning. In these systems, prompt template design is a crucial yet under-explored factor influencing performance. This paper presents a large-scale empirical study to investigate this factor, evaluating 24 different prompt templates on the HotpotQA dataset. The set includes a standard RAG prompt, nine well-formed techniques from the literature, and 14 novel hybrid variants, all tested on two prominent SLMs: Qwen2.5-3B-Instruct and Gemma3-4B-It. Our findings, based on a test set of 18720 instances, reveal significant performance gains of up to 83% on Qwen2.5 and 84.5% on Gemma3-4B-It, yielding an improvement of up to 6% for both models compared to the Standard RAG prompt. This research also offers concrete analysis and actionable recommendations for designing effective and efficient prompts for SLM-based RAG systems, practically for deployment in resource-constrained environments.**

## 1. INTRODUCTION

While large language models have driven major advances in AI, their high computational requirements limit their practical use. Small Language Models (SLMs) offer a more efficient alternative, especially in resource-limited settings. However, SLMs struggle with complex reasoning tasks that require combining information from multiple sources. Retrieval-Augmented Generation (RAG) can help by providing external knowledge, but the success of this method depends heavily on how the model is guided. This paper addresses this central issue, exploring how effective prompt design can improve the performance of SLM-based RAG systems for answering complex questions [1, 2].

## 1.1 Retrieval-Augmented Generation (RAG)

Retrieval-Augmented Generation is a hybrid paradigm that enhances language models by grounding their outputs in external, verifiable knowledge. It directly addresses core language models limitations, such as knowledge cutoffs and hallucination [3].

The typical RAG architecture includes an information retrieval component that, for a given query, retrieves relevant documents from a knowledge base. The retrieved context is added to the language model prompt such that the model can generate a factually accurate and current response [3].

RAG allows models to draw on confidential or up-to-date knowledge without having to be completely retrained, and it is therefore an important method of building reliable AI systems [4, 5].

## 1.2 Small Language Models and Multi-Hop Question Answering

RAG research is currently centered around Small Language Models (SLMs) rather than large-scale models. SLMs have demonstrated significant advancements in computational efficiency and deployment ability. However, the small size of SLM's parameters, restricts their reasoning depth, which is a constraint that is magnified in case of complicated tasks such as multi-hop question answering (QA). To solve this type of problem, collecting information from a range of documents is necessary to construct a well-formed response, which taxes the contextual ability of small models. Multi-hop QA is a type of reasoning tasks where the model must collect information from various sources or passages so that it may come to the answer [6, 7, 8, 9].

Single-hop QA may involve fact extraction from a single sentence or paragraph, whereas multi-hop QA is a more accurate description of natural-world information retrieval tasks, where inference making across multiple sources of evidence is necessary. Such a capability is particularly applicable in areas like scientific research, legal reasoning, and industrial decision-making, where answers are not typically found within one document [10, 11, 12].

Consequently, multi-hop QA provides a stringent test of whether or not a system can accomplish ability and whether it can execute sophisticated inference. The HotpotQA dataset, which is a popular benchmark for the task, is a case in point with its presentation of sophisticated questions that require chaining independent facts. Because SLMs often struggle with complex, multi-step reasoning on their own, their performance in a RAG system depends primarily on the quality of the prompt engineering [13, 14, 15, 16].

## 1.3 The Critical Role of Prompt Engineering

Prompt engineering serves as the critical interface that governs the performance of RAG systems, particularly when using SLMs. As SLMs possess limited reasoning capacity and are highly sensitive to input variations, the prompt must do more than simply present information; it must actively structure and guide the model's reasoning process. This scaffolding involves guiding the model to synthesize disparate facts from retrieved documents, navigate noisy or redundant context, and maintain a coherent chain of thought [17].

Effective prompt design is, therefore, not a trivial optimization but a powerful, cost-effective method for unlocking the latent capabilities of SLMs, directly impacting the accuracy, efficiency, and overall reliability of RAG systems [18].

To systematically evaluate the impact of prompt design, our study begins with a standard RAG prompt template, which we define below in figure 1. This baseline allows us to quantify the improvements achieved by our proposed techniques [19, 20].

```
#standart_contex_aware

Context: {context }
Question: {question}
Answer:
```

**Figure1: Standard RAG Prompt Template**

## 1.4 Contribution and Novelty

This paper presents a large-scale empirical study to address a critical research gap: the optimization of prompt engineering for RAG systems using SLMs. We systematically evaluate 24 distinct prompt templates, including nine established techniques and 14 novel hybrid variants, on the HotpotQA dataset using two state-of-the-art SLMs: Qwen2.5-3B-Instruct and Gemma3-4B-It. Our findings, derived from 18,720 test instances, demonstrate that strategic prompt design can boost accuracy by up to 6% over standard RAG methods [21, 22].

A key novelty is our evaluation framework, which uses a GPT-4o-mini-based judge for nuanced scoring beyond simple string matching. Ultimately, this work provides actionable recommendations for designing effective prompts, enhancing the deployment of RAG systems in resource-constrained environments [23, 24].

The primary contributions of this paper are as follows:

- **A Novel taxonomy of Prompt Strategies:** We design and present a comprehensive taxonomy of 24 prompt templates. This includes 1 Basic RAG prompt template and 9 established baseline strategies from existing literature and 14 novel hybrid templates that we developed, which combine elements of Chain-of-Thought (CoT), self-refinement, and hierarchical synthesis tailored for multi-hop reasoning.

- **A Multi-Dimensional Performance Analysis:** We conduct a rigorous analysis evaluating each prompt across multiple dimensions: answer correctness, processing time, and a composite efficiency metric that balances both factors.

- **Extensive Empirical Results:** We provide a detailed analysis that identifies which prompt engineering techniques yield the best performance for each SLM. Our findings demonstrate that our novel hybrid prompts can improve accuracy by 6% over standard baseline, also, we quantify the specific trade-offs between performance and processing time.

- **Actionable Guidelines for Practitioners:** Based on our empirical results, we distill a set of practical, actionable guidelines to help practitioners select and implement optimal prompt templates for their specific RAG applications, considering both performance goals and resource constraints. All tested prompts are provided in the Appendix.

## 2. RELATED WORK

This section reviews the literature across three key domains relevant to our study: the evolution of RAG architectures, the development of advanced prompt engineering techniques, and the landscape of SLMs and their evaluation benchmarks. We conclude by highlighting the research gaps that serve as the motivation for our work.

## 2.1 Evolution of RAG Architectures

Retrieval-Augmented Generation was first introduced by Lewis et al. [1], who showed that it was possible to realize considerable performance benefit on knowledge-grounded tasks by cross-associating the parametric knowledge within a language model with non-parametric knowledge from an external corpus. Early RAG models developed a straightforward "retrieve-then-read" pipeline, commonly employing dense passage retrievers to obtain relevant documents, which are then used as context to present to the generator model [1].

Follow-up work postulated more advanced architectures, typically in the Advanced RAG, Modular RAG and Agentic RAG categories. They differ from the naive pipeline in possessing more advanced pre-retrieval, retrieval and post-retrieval processing elements. Some of the key enhancements include query transformation, document chunking with intelligence, and re-ranking modules to rank very relevant context [25, 26, 27].

## 2.2 Advanced Prompt Engineering for Complex Reasoning

The effectiveness of the generator of a RAG system is critically dependent on prompt engineering. This field has evolved from simple zero-shot instructions to highly structured reasoning ways. A seminal contribution was Chain-of-Thought (CoT) prompting, which guides a model to break down a problem by generating a sequence of intermediate reasoning steps [17].

This technique has been shown to dramatically improve performance on complex arithmetic and symbolic reasoning tasks. Variations such as Least-to-Most prompting further refine this approach by decomposing a problem into simpler sub-questions that are solved sequentially [22].

Beyond problem decomposition, self-correction and refinement techniques have emerged to enhance reliability. Frameworks like Self-Refine and Chain-of-Verification establish an internal feedback loop where the model generates an initial response, evaluates its own output against a set of criteria, and then iteratively refines it [19].

While powerful in large-scale models, these prompting techniques have been largely unexplored for SLMs. The systematic adaptation and combination of these methods into hybrid prompts for SLMs remains a significant research gap. Existing literature on SLMs has predominantly focused on alternative approaches like query refinement or self-correction, rather than prompt engineering. Our work addresses this gap by conducting a thorough evaluation of prompt strategies for SLMs, specifically targeting complex multi-hop question answering [28].

## 2.3 SLMs, Benchmarks and Evaluation

The growing adoption of SLMs is largely driven by their computational efficiency and also their accessibility. Despite their compact size, state-of-the-art SLMs such as Qwen2.5-3B-Instruct and Gemma3-4B-It have shown strong capabilities in instruction following. However, their limited parameter scale inherently constrains their capacity for deep reasoning over long and complex contexts. Retrieval-Augmented Generation (RAG) has been a main method to combat these obstacles and increase the reasoning capability of SLMs [13, 14].

To compare these systems fairly, testing must use the same set of criteria. The RAGBench dataset provides a large-scale, unified structure for evaluating RAG systems across different domains [8]. Our work focuses on its HotpotQA component, which is the gold standard for multi-hop question answering.

HotpotQA is particularly challenging as its questions are designed to require reasoning across multiple documents, forcing models to identify and connect "bridging" information while ignoring distractors. For evaluation, frameworks such as RAGAS have standardized the measurement of faithfulness and answer correctness [29, 30].

However, these automated metrics often fall short in capturing nuanced aspects of reasoning quality. To address this, recent studies—most notably G-Eval—have popularized the use of powerful LLMs as judges, offering more human-aligned and fine-grained assessments. Building on this methodology, our study adopts an LLM-as-a-judge framework to provide a more comprehensive evaluation of prompt strategies. We also experimented with RAGAS; however, its metrics did not align well with the goals of our study, a limitation we discuss in more detail later [31, 32].

### 2.4 Gaps in Current Research

Despite significant progress in these individual domains, critical research gaps remain at their intersection.

Our review of the literature highlights the following:

- **A Disproportionate Focus on Large Models:** Most research into cutting-edge RAG and prompt engineering has been on large-scale models. The performance behavior, sensitivity, and best prompting strategies of SLMs are thus relatively unexplored and under-discovered.

- **Lack of Systematic Hybrid Prompt Evaluation:** Although numerous varied individual prompting methods have been developed, few systematic investigations have addressed how to design and examine hybrid prompts integrating several methods, especially for the new challenges of multi-hop QA in SLMs for RAG tasks.

- **Prevalence of One-Dimensional Evaluation:** Most research focuses on accuracy of answers, frequently compromising on important trade-offs with computational efficiency. A complete, multi-faceted analysis that simultaneously addresses accuracy, latency, and overall efficiency for RAG-boosted SLMs is mostly missing from the literature.

Our study is therefore designed to directly address these gaps. We conduct a large-scale, multi-dimensional benchmark of both established and novel hybrid prompts, specifically tailored for state-of-the-art SLMs on a challenging multi-hop reasoning QA in RAG tasks [7, 9, 10, 31].

## 3. METHODOLOGY

This section elaborates on the experimental setup designed to evaluate the performance of different prompt engineering strategies in a RAG context. We describe the dataset and data processing steps, the architecture of our pipeline, all the 24 prompt templates under investigation, and the multi-dimensional metrics used for our evaluation.

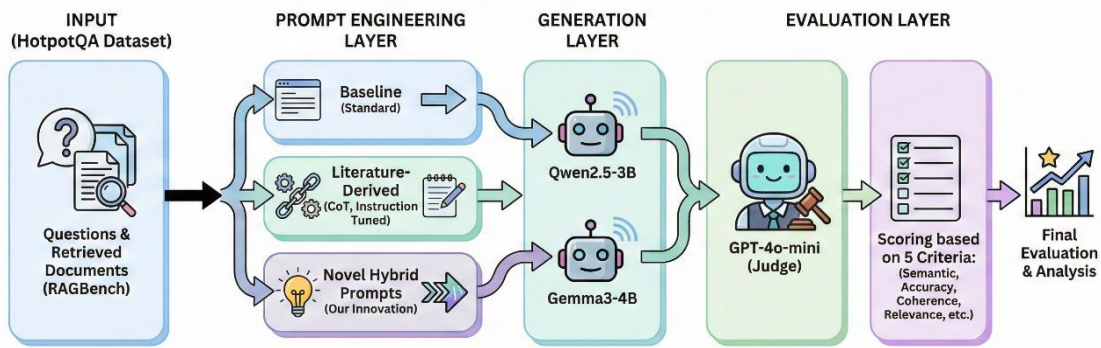

**Figure2: Research Methodology Pipeline**

### 3.1 Dataset and Data Processing

For our experiments, we utilize the test split of the HotpotQA dataset as provided by the RAGBench dataset. HotpotQA is considered the gold standard for multi-hop question answering and is ideal for this study due to several key features:

1. Questions are explicitly designed to require reasoning over multiple supporting documents.

2. Questions are diverse and not constrained by a pre-existing knowledge base.

3. Sentence-level supporting facts are provided, enabling explainable evaluation.

4. The inclusion of comparison questions rigorously tests a model's ability to synthesize and contrast information.

The RAGBench dataset conveniently provides each instance annotated with the question, ground-truth answer, and a set of retrieved documents. The test case of this dataset is a sample of 390 rows [6, 8].

## 3.2 RAG Pipeline Architecture

Our RAG pipeline follows the standard RAG pattern with some changes. The RAGBench dataset is particularly well-suited for our study, as it already provides pre-retrieved passages and distractors, allowing us to focus primarily on prompt design and generation rather than retrieval. Accordingly, our pipeline emphasizes two stages as described below:

- **Generator**: The retrieved documents from the dataset are concatenated to form a single context string. This context, along with the original question, is inserted into a prompt template. The complete prompt is then passed to one of two SLMs to generate an answer: Qwen2.5-3B-Instruct and Gemma3-4B-It.

- **Evaluator**: The generated answer is scored against the ground-truth answer using a separate, powerful large language model, GPT-4o-mini, which serves as an impartial judge. Details of the evaluation methodology are discussed in the following sections.

## 3.3 Prompt Templates

The central part of our study is the systematic evaluation of 24 distinct prompt templates. These templates are categorized into three groups:

- a baseline standard RAG prompt template
- a set derived from existing literature
- and a novel set of hybrid templates designed specifically for this study

The baseline RAG prompt template as the *standard_context_aware* Prompt, shown in Table 1, serves as our primary reference point for comparison against all other prompts. The literature-derived set consists of 9 templates representing established techniques in prompt engineering.

These templates are further grouped into families reflecting foundational, reasoning, self-correction, and advanced strategies.

The novel hybrid prompt set comprises 14 new templates developed by our team. These prompts are specifically designed to tackle the challenges of multi-hop question answering for SLMs by combining elements from the literature families into more robust and guided reasoning structures. We refer to this collection as the Hybrid category.

The identifiers and brief descriptions of all 24 prompt templates are provided in Table 1. Detailed information, including the full prompt bodies can be found in the Appendix C [17, 19, 22].

**Table 1: All Tested Prompt Templates with Descriptions**

| Category | Count | Prompt Templates | Description |
|---|---|---|---|
| Baseline | 1 | standard_context_aware | The simple, standard RAG template |
| Foundational | 2 | instruction_tuned, role_playing | Basic RAG approaches with context presentation |
| Reasoning | 2 | chain_of_thought, least_to_most | Step-by-step thinking and question decomposition |
| Self-correction | 2 | self_refine, chain_of_verification | Draft-critique-refine and claim verification methods |
| Advanced | 3 | few_shot_exemplar, hierarchical_synthesis, structured_json | Examples, citations, and structured outputs |
| Hybrid | 14 | expert_cross_examination_synthesis, adaptive_hybrid, hybrid_focused, hybrid_scientific, hybrid_scientific_synthesis, expert_synthesis, expert_hierarchical_self_refine, optimized_multi_hop, multi_hop_decomposition, optimized_multihop_3b, slm_hotpot_smec, multi_hop_chain, multi_hop_compact, optimized_multi_hop_slm | Novel combinations of multiple techniques |

### 3.4 Evaluation Metrics

To ensure a holistic and rigorous assessment of each prompt, we employ a multifaceted evaluation way. Every generated response is evaluated across three primary dimensions: Answer Quality, Response Latency, and Efficiency.

### 3.4.1 Initial Approach and the Limitations of RAGAS

In a preliminary evaluation phase, we employed the widely adopted RAGAS framework, specifically focusing on its *answer_correctness* metric. This pilot study was carried out on a distinct subset of 5835 question–answer pairs from HotpotQA. However, upon human and team-based verification, the results revealed significant limitations of this approach for our specific use case.

We observed numerous instances where the *answer_correctness* score was unreliably low. For example, generated answers that were semantically and factually correct were frequently assigned scores as low as 0.2 out of 1.0 (see Appendix A for examples).

Our analysis indicated that the metric's heavy reliance on lexical overlap and keyword matching was inadequate for our task. The advanced prompts used in our study are designed to elicit reasoning, and the corresponding answers, while correct, often use different phrasing or have a more extended logical structure than the ground truth. The *answer_correctness* metric usually fails to capture this semantic equivalence and unfairly penalized valid, reasoned responses. This finding motivated our decision to adopt a more robust, semantically-aware evaluation methodology based on the "LLM-as-a-judge" paradigm [29, 30, 31].

### 3.4.2 LLM-based Quality Score ($S_{LLM}$)

To overcome the limitations of RAGAS metrics, we designed and implemented a custom "LLM-as-a-judge" evaluation method. This paradigm is an increasingly accepted methodology in NLP research, as it leverages the nuanced understanding of powerful LLMs to approximate the qualitative assessment of a human expert, but in a scalable and reproducible manner. For this role, we selected GPT-4o-mini as our expert judge model due to its strong reasoning capabilities and cost-effectiveness.

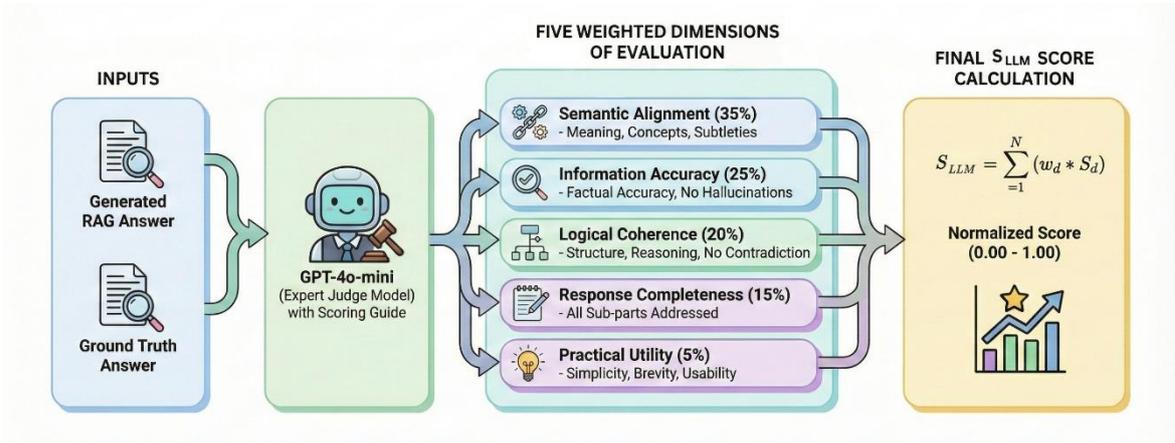

**Figure 3:** S<sub>LLM</sub> Evaluation Pipeline

To ensure a structured and consistent assessment, the judge model was guided by a scoring guide embedded in its prompt (see evaluation prompt in Appendix B). It scores each generated *rag_answer* against the *ground_truth* by evaluating it across five weighted dimensions:

- **Semantic Alignment (35%)**: Scores the extent to which the essential meaning, concepts, and subtleties of the ground truth are represented in the output response.

- **Information Accuracy (25%)**: Scores the factual accuracy of all entities, numbers, and relationships introduced in the response, with penalties for hallucinations.

- **Logical Coherence (20%):** Scores the internal coherence and structural soundness of the response, with rational reasoning and no contradiction.

- **Response Completeness (15%):** Checks whether the entire sub-parts and express requirements of the initial question have been fully answered.

- **Practical Utility (5%):** It checks the general simplicity, brevity, and usability of the response from the end-user's perspective.

The weights were empirically assigned based on the relative importance of each dimension for a reliable RAG system. **Semantic Alignment** and **Information Accuracy** are prioritized (collectively 60%) as they directly reflect the correctness of the answer, while **Practical Utility** is weighted lowest (5%) as it pertains to style rather than substance. For each dimension *d*, the judge assigns a score $S_d$ on a continuous scale of 0.0 to 1.0 (where 1.0 represents perfect performance).

The final S<sub>LLM</sub> is the weighted sum of these scores The final quality score, S<sub>LLM</sub>, is calculated as the weighted sum of the scores assigned to each dimension, normalized to a scale of 0.00 to 1.00.

The score is formally defined as:

$$S_{LLM} = \sum_{\{d \in D\}} w_d \cdot S_d \quad (1)$$

where:

- D is the set of evaluation dimensions {semantic alignment, information accuracy, logical coherence, response completeness and practical utility}

- $w_d$ is the predefined weight for each dimension d.

- $S_d$ is the score (0-1) that the judge assigns to the answer for the dimension d.

This weighted, multi-dimensional metric gives a strong and multi-faceted judgment of the overall quality of every response beyond the mere facts alone, and captures the dimensions that characterize a good and credible answer.

### 3.4.3 Processing Time

This measure estimates generation computational expense. We estimate wall-clock time from the time a formatted prompt is entered to the SLM to the time the full response is generated. Latency measurement of this sort is essential to determine practical deployment feasibility in real-world, user-confronting applications where response time is paramount.

All experiments were performed on a single NVIDIA RTX 4090 GPU with 24 GB VRAM. As each model occupied about 5 GB of VRAM, all three models could be run parallelly on a single GPU.

### 3.4.4 Efficiency Score

To understand the trade-off between answer quality and computational cost, we derive a composite Efficiency score. This metric normalizes the quality score by the time taken to achieve it, defined as:

$$Efficiency = S_{LLM} \div Processing\ Time \quad (2)$$

Efficiency is defined as the ratio of the quality score to latency. The unit of measurement is **accuracy-points per second**. The theoretical range is $(0, \infty)$, where a higher value indicates a model that delivers high-quality answers with minimal latency. A higher efficiency score means the model performs well while using less time, making it a key measure for finding prompt strategies that work effectively and are practical for real-time or time-sensitive tasks.

## 4. EVALUATION AND RESULTS

This section presents a comprehensive empirical analysis of our prompt engineering experiments. We evaluate the performance of a RAG baseline prompt, 9 literature-derived prompts, and 14 novel hybrid prompts on two distinct Small Language Models (SLMs): Qwen2.5-3B-Instruct and Gemma3-4B-It.

The evaluation is centered on three key metrics: accuracy, processing time, and efficiency.

For brevity, the tables presented in this section showcase the top-5 performing prompts for each category. The complete, unabridged results for all tested prompts are available in Appendix E and the prompt body available in Appendix C. Furthermore, more graphical representations and plots from results can be found in Appendix D.

### 4.1 Performance Evaluation on Qwen2.5 3B

Our investigation began by benchmarking the performance of prompt strategies on the Qwen2.5-3B-Instruct model. We systematically compared *standard_context_aware* baseline against both established and novel hybrid prompt templates.

### 4.1.1 Baseline vs. Literature-Derived Prompts

First, we assessed the performance of nine prompts sourced from established literature against our standard baseline. This comparison served to ground our results within the context of existing techniques. The top-performing prompts from this evaluation are shown in Table 2.

### 4.1.2 Baseline vs. Novel Hybrid Prompts

Next, we evaluated our 14 novel hybrid prompts, specifically designed to enhance multi-hop reasoning in SLMs. The results, presented in Table 3, indicate that while these novel prompts successfully elevated the maximum achievable accuracy, they introduced a latency trade-off.

**Table 2: Top 5 Literature-Derived Prompts vs. Baseline for Qwen2.5 3B**

| Rank | Accuracy | | Time | | Efficiency | |
|---|---|---|---|---|---|---|
| | Prompt Method | Avg. Accuracy | Prompt Method | Avg. Time(s) | Prompt Method | Efficiency |
| 1 | hierarchical_synthesis | 0.787 | instruction_tuned | 0.831 | instruction_tuned | 0.931 |
| 2 | structured_json | 0.776 | few_shot_exemplar | 1.037 | few_shot_exemplar | 0.738 |
| 3 | instruction_tuned | 0.774 | Standard_context_aware (baseline) | 1.185 | Standard_context_aware (baseline) | 0.647 |
| 4 | chain_of_thought | 0.772 | structured_json | 1.385 | structured_json | 0.560 |
| 5 | chain_of_verification | 0.770 | role_playing | 1.448 | role_playing | 0.520 |
| Baseline | Ranked 6 | 0.768 | Ranked 3 | 1.185 | Ranked 3 | 0.647 |

**Table 3: Top 5 Novel Hybrid Prompts vs. Baseline for Qwen2.5 3B**

| Rank | Accuracy | | Time | | Efficiency | |
|---|---|---|---|---|---|---|
| | Prompt Method | Avg. Accuracy | Prompt Method | Avg. Time (s) | Prompt Method | Efficiency |
| 1 | optimized_multi_hop_prompting | 0.830 | Standard_context_aware (baseline) | 1.185 | Standard_context_aware(baseline) | 0.647 |
| 2 | multi_hop_compact prompting | 0.825 | slm_hotpot_smec | 9.851 | multi_hop_compact_prompting | 0.081 |
| 3 | multi_hop_decomposition_prompting | 0.806 | multi_hop_decomposition_prompting | 9.951 | multi_hop_decomposition prompting | 0.081 |
| 4 | multi_hop_chain_prompting | 0.801 | multi_hop_compact_prompting | 10.144 | slm_hotpot_smec | 0.076 |
| 5 | hybrid_scientific_prompting | 0.799 | expert_synthesis_prompting | 12.733 | optimized_multi_hop_prompting | 0.061 |
| Baseline | Ranked 11 | 0.768 | Ranked 1 | 1.185 | Ranked 1 | 0.647 |

The results in Table 3 demonstrate that our novel hybrid prompts significantly improve accuracy compared to the baseline. For instance, the top hybrid prompt, *optimized_multi_hop_prompting*, achieved an average accuracy of 83%, whereas the baseline (*Standard_context_aware*) ranks only 11th with 76%.

This shows a clear advantage of the new methods in enhancing response correctness. However, this gain in accuracy comes at a cost: due to increased reasoning tokens and richer analytical content in the responses, the hybrid prompts exhibit higher processing times and lower efficiency scores. The baseline and literature-derived prompts perform better in terms of speed because they generate more concise responses with fewer tokens and do not require the model to output intermediate reasoning steps, significantly reducing computational overhead.

Therefore, the hybrid prompts are most suitable for tasks where accuracy is paramount, while literature-derived prompts are preferable in edge environments where response time and latency efficiency are critical.

The complete set of prompts is provided in Appendix C, and future work could explore improving efficiency by modifying these prompts or instructing the model to provide only the final or more concise answers.

## 4.2 Performance Evaluation on Gemma-3 4B

We replicated the full suite of tests on the Gemma3-4B-It model to assess its capabilities and compare its performance profile against Qwen2.5-3B-Instruct.

### 4.2.1 Baseline vs. Literature-Derived Prompts

Similar to the Qwen2.5-3B-Instruct evaluation, we first compared the baseline against the nine literature-derived prompts to establish a performance benchmark for Gemma3-4B-It. The top results are summarized in Table 4.

### 4.2.2 Baseline vs. Novel Hybrid Prompts

The evaluation of our 14-novel hybrid prompts on Gemma3-4B-It yielded a particularly noteworthy finding, as detailed in Table 6. A key insight emerged where the prompt yielding the highest accuracy, *expert_synthesis_prompting* 84%, was concurrently the fastest (6.98s) and most efficient (0.121) among the entire novel set.

This result suggests that for a sufficiently capable model Gemma3-4B-It, a well-crafted, high-level instruction can be more effective and computationally cheaper than prompts that enforce lengthy, explicit step-by-step reasoning chains. The model was able to internally manage the multi-hop reasoning process efficiently without verbose guidance.

Table 4: Top 5 Literature-Derived Prompts vs. Baseline for Gemma-3 4B

| Rank | Accuracy | | Time | | Efficiency | |
|---|---|---|---|---|---|---|
| | Prompt Method | Avg. Accuracy | Prompt Method | Avg. Time(s) | Prompt Method | Efficiency |
| 1 | hierarchical_ synthesis | 0.842 | instruction_tuned | 0.510 | instruction_tuned | 1.412 |
| 2 | role_playing | 0.833 | Standard_context_ aware(baseline) | 0.709 | Standard_context _aware(baseline) | 1.107 |
| 3 | chain_of_ thought | 0.831 | few_shot_exemplar | 0.792 | few_shot_ exemplar | 1.028 |
| 4 | chain_of_ verification | 0.824 | hierarchical_synth esis | 1.207 | hierarchical_ synthesis | 0.698 |
| 5 | least_to_most | 0.817 | role_playing | 1.288 | role_playing | 0.646 |
| Baseline | Ranked 8 | 0.785 | Ranked 2 | 0.709 | Ranked 2 | 1.107 |

Statistics shown in Table 5 reveal that the performance of the Gemma3-4B-It model is better than that of the Qwen2.5-3B-Instruct model in processing time, efficiency, and accuracy. The latter's better performance is partly due to its more numerous parameters (4B versus 3B), which supports the fact that increasing the number of parameters in statistical language models (SLMs) results in boosting accuracy and processing speed. Also, Gemma's custom architecture, designed by Google, is responsible for its powerful performance and was quite significant at its unveiling.

With regard to prompt performance, *instruction_tuned* achieves high efficiency and is thus not inappropriate for latency-sensitive systems, and *hierarchical_synthesis* achieves optimal accuracy and is therefore preferred for correctness-critical tasks.

These results represent the evaluation of the baseline plus 9 literature-derived prompts. The following section will present the results for our 14 novel hybrid prompts, which are expected to further improve accuracy for environments where precision is critical and processing time is less of a concern.

**Table 5: Top 5 Novel Hybrid Prompts vs. Baseline for Gemma-3 4B**

| Rank | Accuracy | | Time | | Efficiency | |
|---|---|---|---|---|---|---|
| | Prompt Method | Avg. Accuracy | Prompt Method | Avg. Time(s) | Prompt Method | Efficiency |
| 1 | expert_synthesis_ prompting | 0.846 | Standard_context_a war (baseline) | 0.709 | Standard_context_a ware(baseline) | 1.107 |
| 2 | multi_hop_compact _prompting | 0.841 | expert_synthesis_ prompting | 6.98 | expert_synthesis_ prompting | 0.121 |
| 3 | multi_hop_ decomposition_ prompting | 0.832 | slm_hotpot_smec | 7.08 | slm_hotpot_smec | 0.110 |
| 4 | optimized_multihop _3b_prompting | 0.831 | multi_hop_ decomposition_ prompting | 8.01 | multi_hop_ decomposition_ prompting | 0.104 |
| 5 | optimized_multi_ho p_prompting | 0.809 | multi_hop_compact _prompting | 8.37 | multi_hop_compact _ prompting | 0.100 |
| Baseline | Ranked 8 | 0.785 | Ranked 1 | 0.709 | Ranked 1 | 1.107 |

## 4.3 Cross-Model Performance Synthesis

A comprehensive analysis across both models reveals a critical trade-off between accuracy and computational efficiency, defining two distinct use-cases for different prompt categories.

- **Superior Baseline Performance of Gemma3:** Firstly, Gemma3-4B-It consistently outperforms Qwen2.5-3B-Instruct across all three metrics. Its larger size and advanced architecture provide a stronger foundation, resulting in higher accuracy, lower latency, and greater efficiency, regardless of the prompt strategy employed.

- **Literature-Derived Prompts for Speed and Efficiency:** On both models, literature-derived prompts like *instruction_tuned* and *few_shot_exemplar* are undisputed champions of speed and efficiency. Their simplified architecture makes for lightweight processing times, making them eminently well suited to those applications requiring real-time output, those deployments on edge hardware, and those use cases for which minimal latency is at a premium. As good as they are on accuracy, they do not reach the optimal performance levels that we see with our novel approaches.

- **Novel Hybrid Prompts for Peak Accuracy:** Our 14 novel hybrid prompts were created to have maximum depth of reasoning, and the results bear witness to their effectiveness in that direction. The accuracy ceiling is consistently set on both models by these, and performance is improved beyond that possible using normal methods. This is at the price of substantially increased processing times. This renders them optimal for special-purpose, high-risk application fields like medical diagnosis, scientific discovery, or financial forecasting, where precision is of overriding importance and a more prolonged response time is an acceptable compromise.

Briefly, the optimal choice of strategy is not a one-size-fits-all answers but is dictated by the specific needs of the problem. For general-purpose, time-critical implementations, classic literature prompts provide optimal performance. For special-purpose, error-sensitive implementations, our new hybrid prompts provide an efficient, but more resource-intensive, alternative.

# 5. DISCUSSION AND ANALYSIS

The results of our comprehensive evaluation provide several key insights into the role of prompt engineering for SLM-based RAG systems. This section analyzes these findings, discusses their implications, and offers actionable recommendations based on the empirical data from Section 4.

## 5.1 The Accuracy-Latency Frontier: A Steep but Nuanced Trade-Off

The key result of this work uncovers a significant accuracy-latency trade-off. The novel hybrid prompts significantly improved the optimum accuracy at a significant computational cost. Our results showed a rise in accuracy from baseline of 76.8% to a maximum of 83.0% on Qwen2.5 3B, which is a 6.2% increase, and from 78.5% to 84.6% on Gemma3 4B, which is a 6.1% increase. However, such an improvement came at the cost of a processing time increase by a factor of 8 to 10 compared to baseline. For instance, on Qwen2.5 3B, for optimized performance of the novel prompts, the processing latency jumped from a baseline value of about 1.2 seconds to a value of about 10 seconds.

This establishes a clear accuracy-latency frontier: achieving the highest echelon of reasoning quality with SLMs requires complex, verbose prompts that are inherently slow. However, our results also reveal a crucial nuance: this trade-off is not absolute. The performance of *expert_synthesis_prompting* on Gemma3 4B demonstrates that for more capable models, it is possible to achieve both top-tier accuracy and high efficiency simultaneously among advanced prompts, a point we will explore further.

## 5.2 Model-Specific Affinities: The Primacy of Model Capacity and Architecture

Though design of prompt plays a role, our results affirm architecture and model size as leading causes of performance. Gemma3 4B was more performant than Qwen2.5-3B-Instruct on nearly every measure—accuracy, speed, and efficiency—even on similar prompts. The baseline of Gemma was faster (0.71s compared to 1.19s) and more accurate (78.5% compared to 76.8%) than Qwen's, which underscores its better foundational performance.

Furthermore, optimal prompt varied from model to model, showing model-specific preference and denying "plug-and-go" usage.

- For Qwen2.5 3B, a Balanced Compromise is Key. *The optimized_multi_hop_prompting* template achieved the highest accuracy (83.0%), but at a significant latency cost. A more practical balance is struck by *multi_hop_compact_prompting* and *multi_hop_decomposition_prompting*. These prompts delivered nearly top-tier accuracy (82.5% and 80.6%, respectively) while keeping latency manageable (around 10 seconds). They provide enough structure to guide the smaller model without overwhelming it, making them the most efficient high-accuracy options for Qwen2.5.

- For Gemma-3 4B, expert_synthesis_prompting is the Unrivaled Winner. The expert_synthesis_prompting template was uniquely effective, achieving the highest accuracy (84.6%) while also being the fastest (6.98s) and most efficient (0.121) among all novel prompts for Gemma. This remarkable result suggests that Gemma's larger capacity and advanced architecture allow it to effectively interpret a high-level "expert synthesis" instruction. It can harness its internal reasoning power without the need for the verbose, explicit hand-holding required by more granular Chain-of-Thought style prompts, thus breaking the typical accuracy-latency trade-off seen elsewhere.

## 5.3 Analysis of Novel Prompt Strategies

The performance of different hybrid categories provides further insights.

- Decomposition as a Robust Strategy: Prompts like *multi_hop_compact_prompting* and *multi_hop_decomposition_prompting* were highly effective on both models. This confirms that breaking a complex question into smaller, manageable parts is a robust strategy for improving SLM reasoning, particularly for smaller models like Qwen2.5 3B.

- Ineffectiveness of Simple Constraints: The *slm_hotpot_smec_prompting* prompt, which aimed to improve speed by constraining output, yielded only marginal gains. While it was among the fastest hybrid prompts, its accuracy was lower. This suggests that simply limiting reasoning length is less effective than providing better reasoning structures.

- Power of High-Level Guidance for Capable Models: The standout success of expert_synthesis_prompting on Gemma indicates that for more capable SLMs, framing the task with a strong persona and a clear, high-level goal can be more efficient than micromanaging the reasoning process with explicit, step-by-step instructions.

## 5.4 Practical Recommendations for Practitioners

Our comprehensive analysis summarized as the following actionable recommendations in Table 6. The choice of prompt should be dictated by the specific application requirements and the underlying model.

**Table 6: Recommended Prompt Strategies Based on Task Goal and Model**

| Goal | Recommended for Qwen2.5 3B | Recommended for Gemma3 4B | Description |
|---|---|---|---|
| Maximum Accuracy | optimized_multi_hop_prompting | expert_synthesis_ prompting | Prioritizes quality above all else suitable for offline, non-interactive tasks. |
| Maximum Efficiency | instruction_tuned | instruction_tuned | Delivers the fastest response times which is ideal for real-time chatbots and also latency-critical applications. |
| Best Overall Balance | multi_hop_compact_ prompting | expert_synthesis_ prompting | Offers the best trade-off, achieving top-tier accuracy with more manageable latency for the respective model. |

## 5.5 Key Findings

The present large-scale empirical study on prompt engineering for SLM-based RAG systems yielded several critical findings:

1. A Steep Accuracy-Efficiency Trade-Off can be seen, but It Is Not Absolute. Increasing prompt complexity to improve reasoning consistently leads to higher accuracy at the cost of a significant (8-10x) increase in processing time. However, the case of *expert_synthesis_prompting* on Gemma3 4B shows that highly capable models can break this trend, achieving peak accuracy and high efficiency simultaneously among advanced prompts.

2. Simplicity is King for Ultimate Efficiency. On both models, the short, literature-based instruction_tuned prompt was overwhelmingly the efficiency champion. The short instructions had the smallest latency and were optimal for real-time usage, where responsiveness matters most.

3. Synthesis-Based Prompts Achieve Peak Accuracy. Prompts that explicitly instruct the model to decompose, integrate, and synthesize information from multiple documents—such as *optimized_multi_hop_prompting* and *expert_synthesis_prompting*, consistently delivered the highest accuracy scores.

4. There is No "Best" Prompt: Model-Specific Affinity is Crucial. The performance of a prompt is highly dependent on the underlying SLM. The optimal prompt for achieving accuracy, efficiency, or a balance between the two was different for Qwen2.5 and Gemma-3, invalidating a one-size-fits-all approach.

5.  Model Scale and Architecture are Primary Drivers of Performance. While prompt engineering is a critical optimization lever, the fundamental capabilities of the model remain paramount. The larger Gemma3-4B-It model consistently outperformed the Qwen2.5-3B-Instruct model, delivering higher accuracy, lower latency, and greater overall efficiency across the vast majority of prompts.

# 6. CONCLUSION

This paper addressed the critical research gap in finding the best prompt way for Retrieval-Augmented Generation (RAG) for Small Language Models (SLMs) in the context of complex multi-hop question answering. Through a large-scale empirical study evaluating 24 distinct prompt templates on Qwen2.5-3B-Instruct and Gemma3 4B, our multi-dimensional analysis elucidated the intricate interplay between answer accuracy, processing latency, and overall efficiency.

Our findings provide conclusive evidence that prompt engineering is a decisive factor in system performance. We demonstrated that carefully designed hybrid prompts can achieve substantial accuracy gains, pushing performance to peak scores of 83.0% on Qwen2.5-3B-Instruct and 84.6% on Gemma3 4B, significantly outperforming standard baseline.

Two central themes emerged from our analysis: a steep but nuanced accuracy-efficiency trade-off and a strong model-specific prompt affinity. While novel hybrid templates always yielded the best degree of precision, it was at the price of an 8-to-10 times increase in latency. By way of comparison, efficient and direct templates such as instruction_tuned provided unparalleled efficiency and were thereby highly favorable for resource-starved deployments. Critically, we identified an important exception to this trade-off with the expert_synthesis_prompting on Gemma3 4B, which achieved both the highest accuracy and the best efficiency among all novel prompts, highlighting that more advanced models can efficiently handle high-level, goal-oriented instructions.

In conclusion, this research provides a detailed, data-driven roadmap for designing and selecting effective prompts. By quantifying the trade-offs and revealing model-specific behaviors, we offer a practical guide for practitioners to unlock the full potential of SLM-based RAG systems in complex, knowledge-intensive applications, enabling them to make informed decisions based on their specific accuracy and latency requirements.

# Appendix for Evaluating Prompt Engineering for RAG in Small Language Models: A Multi-Hop QA Approach

## Appendix A: Comparison of RAGAS' Incorrect Responses with Our Evaluation Outputs

The following examples highlight instances in which the RAGAS system produced incorrect answers according to the *answer_correctness* metric. These cases provided valuable insights into the common failure patterns of RAGAS and informed the development of a novel custom evaluation way for our study.

<p style="text-align:center;color:red;"><b>Example 1 (False Negative)</b></p>

```
Question:
Philip Despencer's brother was a favorite of which king who was deposed in January
of 1327?

Ground truth:
Philip Despencer's brother, Hugh Despenser, the Younger, was a favorite of King
Edward II, who was deposed in January of 1327.

Rag Answer:
Analytical Framework

Step 1 - Evidence Extraction:
From the provided context:
Philip Despencer was born in Stoke, Gloucester around 1290.
He married Margaret de Goushill and had a brother named Hugh Despenser, the
Younger.
Hugh Despenser, the Younger, was described as "a favorite of King Edward II."
Edward II (also called Edward of Caernarfon) was King of England from 1307 until he
was deposed in January 1327.

Step 2 - Mechanistic Reasoning:
Edward II became king following the death of his older brother Alphonso, and Hugh
Despenser, the Younger's role as a favorite would suggest close personal or
political relationships with Edward. Being a "favorite" typically implies
significant influence over the king, often involving important positions in
government.

Step 3 - Critical Assessment:
The evidence is clear and direct, making it reliable. However, there are no
additional details about other potential favorites of King Edward II or any
conflicts that led to his deposition that could be independently verified from this
information alone. The timeframe (1307-1327) for Hugh Despenser's role as a
favorite aligns well with the period before Edward II was deposed in 1327,
suggesting a strong causal link.

Step 4 - Creative Synthesis:
While the evidence points to Edward II as the king whose deposition we are looking
at, it is important to consider broader historical context. The relationship
between Hugh Despenser and King Edward II can be seen as symptomatic of broader
issues within English royal power dynamics in that era. Hugh's influence likely
intensified during a period when tensions were high with baronial opposition
(particularly against the unpopular Earl of Lancaster).

Step 5 - Expert Conclusion:
Based on this analysis, Philip Despencer's brother, Hugh Despenser, the Younger,
was indeed a favorite of King Edward II. Edward II was deposed in January 1327,
making him the king whose deposition we are examining.

Therefore, my definitive answer is:
Hugh Despenser, the Younger's relationship with Philip Despencer suggests that he
was a favorite of King Edward II, who was deposed in January 1327

RAGAS answer_correctness score:
0.378 (😕)

Our custom evaluation score:
0.85 (😊)
```

**Example 2 (False Positive)**

Question:
The people for whom pap is a staple food speak a language that is a major branch of what language family?

Ground truth:
The people for whom pap is a staple food speak a language that is a major branch of the Bantu language family.

Rag Answer:
EVIDENCE EXAMINATION:

[DOC-1]: This document defines "staple food" but does not mention any language family. It is irrelevant to our question about which language family the people who consume pap speak.

[DOC-2]: Provides information about the Apurinã indigenous communities and their endangered status due to low transmission of their language, Apurinã. However, it does not explicitly state what language family Apurinã belongs to or if pap is a staple food for that specific group.

[DOC-3]: Discusses Bantu peoples but only in the context of Africa's linguistic landscape and provides an estimate of about 650 Bantu languages. It does not mention any connection between Bantu-speaking people and the consumption of pap.

[DOC-4]: Specifically mentions "Pap" as a traditional porridge made from mielie-meal (ground maize) and a staple food in Southern Africa, belonging to the Bantu peoples. However, it does not address which language family these Bantu-speaking people belong to or if Pap is considered a staple for them.

CROSS-EXAMINATION:
[DOC-4] explicitly mentions that pap is a traditional porridge made from mielie-meal (ground maize) and describes its consumption by the Bantu peoples of Southern Africa. While [DOC-3] provides information about the Bantu language family, it does not specify if this information applies to pap consumption or if Apurinã has any relation to the Bantu-speaking populations.

[DOC-2], on the other hand, focuses entirely on the Apurinã people and their endangered status. It discusses their unique syntax (active-stative) but does not provide direct evidence linking them to either Bantu languages or pap consumption.

Synthesizing this information:
The documents [DOC-4] and [DOC-3] clearly indicate that pap is a staple food in Southern Africa and belongs to the Bantu-speaking population. However, there are no explicit links provided between the Apurinã people (as mentioned in [DOC-2]) and either the consumption of pap or the broader Bantu language family.

VERDICT:
Based on the available evidence from [DOC-3] and [DOC-4], which both confirm that pap is a staple food for Bantu-speaking populations, we cannot definitively state that the Apurinã people consume pap. Therefore, while it seems plausible that the general Bantu language family might be relevant to the broader context of Southern African staples, specific evidence linking the Apurinã to the consumption of pap or their linguistic affiliation is lacking.

The answer based on current available documents: The language family for Bantu-speaking populations who consume pap remains uncertain as direct evidence linking Apurinã specifically to these staples and their corresponding languages is not provided.

RAGAS answer_correctness score:
0.91 (😊)

Our custom evaluation score:
0.45 (😕)

# Appendix B: Prompt template used for evaluation, including the LLM-as-a-judge setup

Below, we provide the exact prompt template used in our evaluation process. This template represents the LLM-as-a-judge setup, which guided the assessment of correctness and reasoning in generated answers. By reviewing this, readers can clearly see how the evaluation system was structured and applied in our experiments.

**Custom Evaluation Prompt Template**

```
You are an expert evaluator assessing RAG system outputs across different prompting
strategies.

EVALUATION FRAMEWORK:
Assess the quality of the RAG Answer compared to the Ground Truth, considering both
semantic meaning and practical utility.

SCORING DIMENSIONS (weights in parentheses):

1. SEMANTIC_ALIGNMENT (35%): How well does the answer capture the essential meaning
and concepts?
    - Consider conceptual completeness, not word-for-word matching
    - Evaluate understanding depth and nuance preservation

2. INFORMATION_ACCURACY (25%): Are facts, entities, numbers, and relationships
correct?
    - Verify specific claims against ground truth
    - Check for hallucinations or fabrications

3. LOGICAL_COHERENCE (20%): Is the answer internally consistent and well-
structured?
    - Assess reasoning flow and argument structure
    - Check for contradictions or gaps in logic

4. RESPONSE_COMPLETENESS (15%): Does the answer fully address the question?
    - Evaluate coverage of all question aspects
    - Consider whether key points are missing

5. PRACTICAL_UTILITY (5%): How useful would this answer be to a real user?
    - Consider clarity and actionability
    - Assess if the response provides value

SCORING SCALE:
- Use values between 0.001 and 0.999 (never exactly 0 or 1)
- 0.850-0.999: Excellent performance
- 0.650-0.849: Good performance
- 0.450-0.649: Moderate performance
- 0.250-0.449: Below average performance
- 0.001-0.249: Poor performance

EVALUATION PROCESS:
1. First, identify the core information units in the Ground Truth
2. Map these against what's provided in the RAG Answer
3. Assess each dimension independently
4. Calculate weighted final score
5. Create Final Response = SEMANTIC_ALIGNMENT + INFORMATION_ACCURACY +
LOGICAL_COHERENCE + RESPONSE_COMPLETENESS + PRACTICAL_UTILITY (Result will be from
0.000 to 1.000)

OUTPUT REQUIREMENT:
Your evaluation should result in a single floating-point number between 0.000
(completely incorrect) and 1.000 (perfectly correct).
Do not provide any explanation or additional text. Your response must be only the
numeric score with two decimal places.

QUESTION:
{data['question']}
-----------------------------
GROUND TRUTH:
{data['ground_truth']}
-----------------------------
RAG ANSWER:
{data['rag_answer']}
-----------------------------

Your Numeric Evaluation (0.000 to 1.000):
```

# Appendix C: Full templates of all 24 prompts discussed in the paper

This appendix provides the complete text and Description for all 24 prompt templates evaluated in this study. The {context} and {question} placeholders are dynamically replaced with the corresponding data for each instance during the experiments.

## C.1) baseline + 9 prompts from literature:

```
standard_context_aware (baseline)

Context: {context}
Question: {question}
Answer:
```

```
instruction_tuned

Use the provided context to answer the question. Do not use any other knowledge.

Context: {context}
Question: {question}
Answer:
```

```
role_playing

You are an expert researcher and analyst. Based ONLY on the retrieved documents
below, explain the answer to the following question.

Context: {context}
Question: {question}
Answer:
```

```
chain_of_thought

Based on the context, answer the following question. First, think step by step to
break down your reasoning. Then, provide the final answer.

Context: {context}
Question: {question}
Answer:
Step-by-step reasoning:
1.
2.

Final Answer:
```

```
least_to_most

To answer the main question, we must first answer a series of sub-questions based
on the provided context.

Context: {context}
Question: {question}

Decompose the question into sub-questions and answer them.
Sub-question 1: [Generate and answer sub-question 1]
Sub-question 2: [Generate and answer sub-question 2]
…

Therefore, the final answer to the main question is:
```

```
self_refine

Your task is to answer the question based on the context in a two-step process.

Context: {context}
Question: {question}

First, produce a draft answer.
Draft Answer:

Second, critique the draft answer for accuracy and completeness based on the
context. Then, provide the final, refined answer.
Refined Answer:
```

```
chain_of_verification

Answer the following question based on the context.
1. First, generate a baseline answer.
2. Next, break down the baseline answer into individual verification-worthy claims.
3. For each claim, check if it is supported by the context.
4. Finally, generate a final, verified answer based on the verified claims.

Context: {context}
Question: {question}
Response:
```

```
few_shot_examplar

Here are some examples of how to answer questions based on context:

Example 1:
Context: The study shows that exercise improves cognitive function.
Question: What does exercise do to cognitive function?
Answer: According to the context, exercise improves cognitive function.

Example 2:
Context: Machine learning models require large datasets for training.
Question: What do ML models need for training?
Answer: Based on the context, machine learning models require large datasets for
training.

Now answer this question:
Context: {context}
Question: {question}
Answer:
```

```
hierarchical_synthesis

You have been provided with multiple documents. Your task is to synthesize the
information from these documents to answer the question. When you use information
from a document, cite it using its identifier.

Context:
[DOC-1] {context_part_1}
[DOC-2] {context_part_2}
...
Question: {question}
Answer (with citations):
```

```
structured_json

Based on the context, extract the required information and format it as a JSON
object. The JSON object must have the following keys: "answer", "confidence_score",
"supporting_evidence".

Context: {context}
Question: {question}
JSON Response:
```

**C.2) Our 14 Novel Hybrid Prompts:**

```
expert_cross_examination_synthesis_prompting

You are an expert researcher and analyst conducting a rigorous cross-examination of
evidence to answer research questions. Always reference specific documents using
their identifiers [DOC-X] when making any claim.

Examples of Expert Cross-Examination:
... (Examples as provided in the source code) ...

Now conduct your expert cross-examination:

Documents:
[DOC-1] {context_part_1}
...
Question: {question}

Expert Cross-Examination:
Evidence Examination:
[Examine each document critically and cite using [DOC-X] format]

Cross-Examination:
[Identify contradictions, assess reliability, weigh evidence strength - always cite
sources]

Synthesis:
[Integrate information across all documents with proper citations, showing how they
connect or conflict]

Verdict:
[Final conclusion based on synthesized evidence with clear document citations]
```

```
hybrid_focused_prompting

You are an expert researcher conducting an in-depth analysis. Demonstrate both
analytical rigor and creative scientific thinking.

Context: {context}
Question: {question}
Analytical Framework:
Step 1 - Evidence Extraction:
What specific facts, data, and findings are presented?

Step 2 - Mechanistic Reasoning:
What underlying processes or principles explain these findings?

Step 3 - Critical Assessment:
How reliable and complete is this evidence? What are the limitations?

Step 4 - Creative Synthesis:
What novel insights, connections, or implications can be drawn?

Step 5 - Expert Conclusion:
Based on this analysis, what is your definitive, evidence-based answer?

Provide your response following this framework, showing your reasoning process
while delivering a clear, scientifically sound conclusion:
```

```
hybrid_scientific_prompting

You are a senior research scientist known for rigorous analysis and creative
insights. Your task is to provide a comprehensive, evidence-based response that
demonstrates both analytical depth and scientific creativity
{context_instruction}

Research Context:
{formatted_contexts}

Research Question: {question}

Analysis Approach:
Follow this systematic framework to ensure thorough analysis:

1. Initial Assessment: What are the key findings and data points?
2. Pattern Recognition: What relationships, trends, or mechanisms can be
identified?
3. Cross-Source Synthesis: (if multiple sources): How do findings complement,
contradict, or extend each other?
4. Critical Evaluation: What are the strengths and limitations of the evidence?
5. Creative Integration: What novel insights or unexpected connections emerge?
6. Scientific Conclusion: What definitive answer can be supported by the evidence?

Expert Analysis:
```

```
hybrid_scientific_synthesis_prompting

You are an expert scientific researcher and analyst, specializing in evidence-based
synthesis. Your role is to answer questions ONLY based on the provided documents,
without external knowledge. Use a multi-step process:

1. Summarize each document briefly, highlighting key facts.
2. Use few-shot examples to guide your reasoning:
... (Examples as provided in source code) ...
3. Break down the question step-by-step (Chain-of-Thought): Identify sub-questions,
answer them from documents, and cite [DOC-X].
4. Draft an initial answer.
5. Self-refine: Critique the draft for accuracy, completeness, and potential biases
based on documents. If insufficient evidence, state 'Insufficient evidence' and
suggest refinements.
6. Provide the final synthesized answer with citations.

Context:
[DOC-1] {context_part_1}
...

Question: {question}

Document Summaries:
[DOC-1]: ...
[DOC-2]: ...

Step-by-Step Reasoning (with sub-questions):

Draft Answer:

Self-Critique:

Final Answer (with citations):
```

```
expert_synthesis_prompting

You are a meticulous and expert research assistant. Your primary goal is to provide
accurate, comprehensive, and evidence-based answers by synthesizing information
from the provided documents.

Core Principles:
1.  Synthesize, Don't Summarize...
2.  Cite Everything ...
3.  Address Contradictions ...
4.  Direct Answer First ...

---
Example Task:
... (Example as provided in source code) ...
---
Your Task:
Provided Documents:
[DOC-1] {context_part_1}
...
Question: {question}

Answer:
```

```
expert and precise analyst. Use ONLY the documents below.
Rules:
- For every claim, cite its source immediately using [DOC-i].
- If the context is insufficient, output "INSUFFICIENT_CONTEXT".
- Do not use outside knowledge. Do not invent claims. Prefer short direct quotes.

Context:
[DOC-1] {context_part_1}
...
Question: {question}

Your output must follow this structure:

Draft (expert's initial answer):
- Answer (2–4 sentences, with [DOC-i] citations):
<initial expert answer with citations>
- Key evidence (2–4 short quotes):
- "..." [DOC-?]
- "..." [DOC-?]

Self-Review (same expert reviews their draft):
- Gaps/Risks (max 3 items; for each, specify which part of the draft and which
document(s) are relevant):
1) <gap/risk> — related to: <draft section> — evidence: [DOC-?]
...
- Fixes (max 3 items; specify exactly what to add/remove and which document
supports it):
1) <fix action> — supported by: [DOC-?]
...

Final (refined version after applying fixes):
- Answer (3–6 sentences, precise and concise, with [DOC-i] citation on each factual
claim):
<final expert answer with citations>
- Confidence: <High|Medium|Low> — one-line justification based on the documents
[DOC-?]
```

optimized_multi_hop_prompting

You are a precision-focused answer engine designed for multi-hop reasoning and rigorous evidence citation.

Critical Instructions:
1) Identify the 2-3 most essential documents needed for this multi-hop question ... (and 7 other instructions as provided in source code) ...

Documents:
[DOC-1] {context_part_1}
...

Question: {question}

Thinking Process:
Step 1: Document selection - ...
Step 2: Fact extraction - ...
Step 3: Reasoning linkage - ...
Step 4: Synthesis - ...
Step 5: Validation - ...

Explanation:
1) Document Selection:
2) Key Facts Extracted:
3) Multi-hop Reasoning:
4) Synthetic Integration:

Answer (must include all facts with [DOC-X] citations):

Multi_hop_decomposition_prompting

You are an expert in multi-hop reasoning.
You have the following documents:
[DOC-1] {context_part_1}
...

Main Question:
{question}

Instructions:
1. Decompose: First, break the main question into clear sub-questions.
2. Answer Sub-Questions: For each sub-question, find the answer in the documents and cite every fact as [DOC-x].
3. Synthesize: Combine the sub-answers into one coherent final answer. Cite the documents that support each part.
4. Output Format:
---
Sub-questions:
1) ...
2) ...

Sub-answers:
1) ... [DOC-?]
2) ... [DOC-?]

Final Answer:
...
---
Be concise, precise, and never invent facts. Rely strictly on the provided documents.

```
optimized_multihop_3b_prompting

Answer the question using ONLY the provided documents. For questions needing
information from multiple documents, work step-by-step.

Multi-Hop Examples:
... (Examples as provided in source code) ...

Your  Task:

Documents:
[DOC-1] {context_part_1}
...

Question: {question}

Step-by-Step-Reasoning:
Step 1: What information do I need to answer this question?
Step 2: Find relevant facts from each document:
Step 3: How do these facts connect to answer the question?
Step 4: Combine the facts for the final answer:

Final Answer: [Provide clear answer with [DOC-X] citations]
```

```
slm_hotpot_smec

You are an evidence-bound assistant. Use ONLY the Documents below. No outside
knowledge.
Task: For the multi-hop question, extract the minimum bridging facts and return a
short, normalized final answer.

Output (exactly these sections):
Findings:
- <fact> — "<short quote>" [D#]
...
Reasoning: <one short sentence combining the findings>
Final_Answer: <only the final answer, normalized>

... (Normalization rules and Constraints as provided in source code) ...

Documents:
[D1] {context_part_1}
...

Question: {question}

Begin.
```

```
multi_hop_chain_prompting

Documents:
Doc1: {context_part_1}
...

Question: {question}

Find the answer by connecting information:

First fact: (What relevant fact from Doc1 or Doc2?)
→ Leads to: (What does this tell us?)
→ Combined with: (What other fact connects?)
→ Therefore: (Final answer)

Answer with citations [Doc1], [Doc2]:
```

```
multi_hop_compact_prompting

Answer the question using ONLY the given documents. Connect information across
documents.

Documents:
[1] {context_part_1}
...

Question: {question}

Instructions:
1. Find relevant facts in each document
2. Connect facts to answer the question
3. Cite sources using [1], [2], etc.

Step 1 - Key facts:
• From [1]:
• From [2]:
(continue as needed)

Step 2 - Connect facts:
(Show how facts relate)

Answer: (Complete answer with citations)
```

```
optimized_multi_hop_slm

You are an expert analyst answering multi-hop questions ONLY from the provided
documents. Cite every fact with [DOC-X]. If evidence is missing, say 'Insufficient
evidence'.

Quick Examples (Learn from these):
... (Examples as provided in source code) ...

Your Task - Follow These Steps Exactly:
1. Break Down Question: ...
2. Synthesize Evidence: ...
3. Draft Answer: ...
4. Quick Self-Refine: ...
5. Final Answer: ...

Documents:
[DOC-1] {context_part_1}
...

Question: {question}

Your Response (use the steps):
```

**Appendix D: Selected images from the evaluations in Section 4 that may be useful**

Here we provide a set of selected images from Section 4 evaluations to help you better understand the results.

**D.1) Qwen2.5 3B, baseline + 9 prompts from literature:**

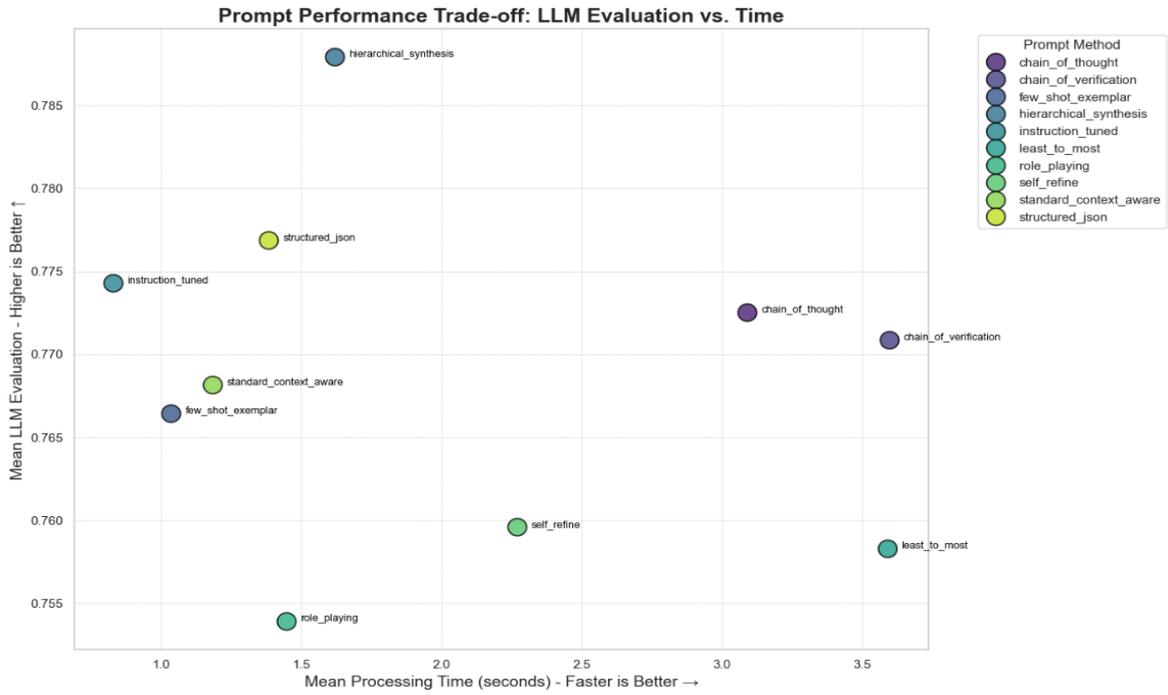

**D.2) Qwen2.5 3B, baseline + 14 novel prompts:**

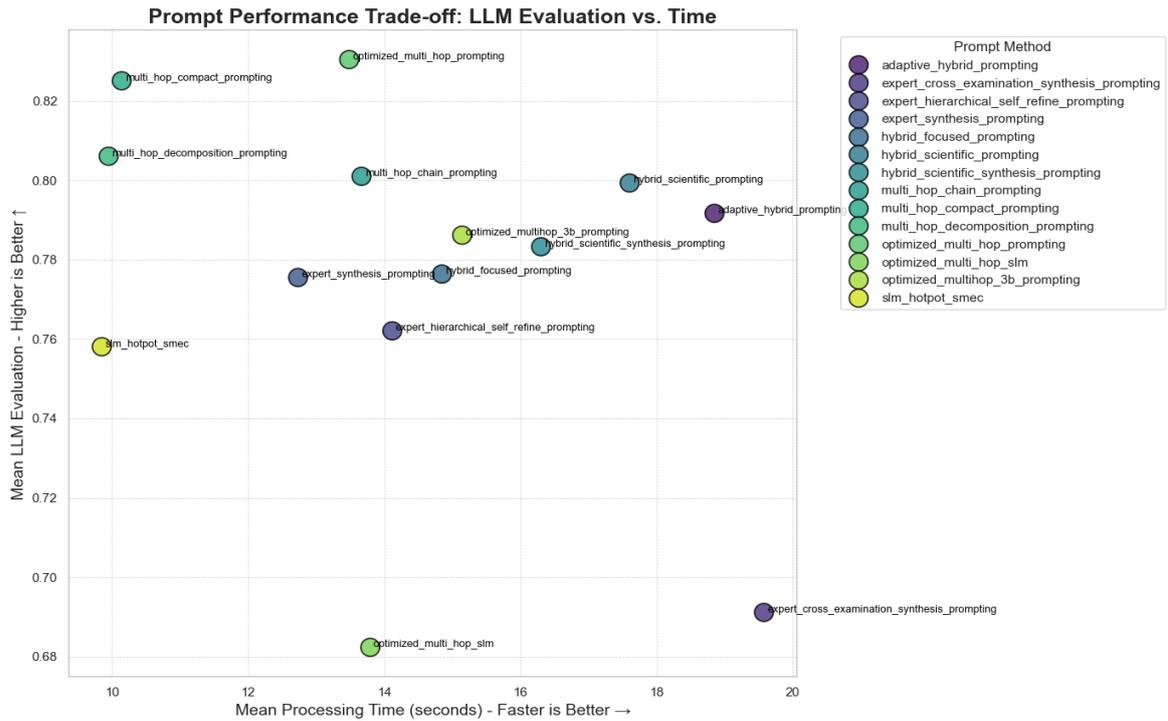

**D.3) Gemma3 4B, baseline + 9 prompts from literature:**

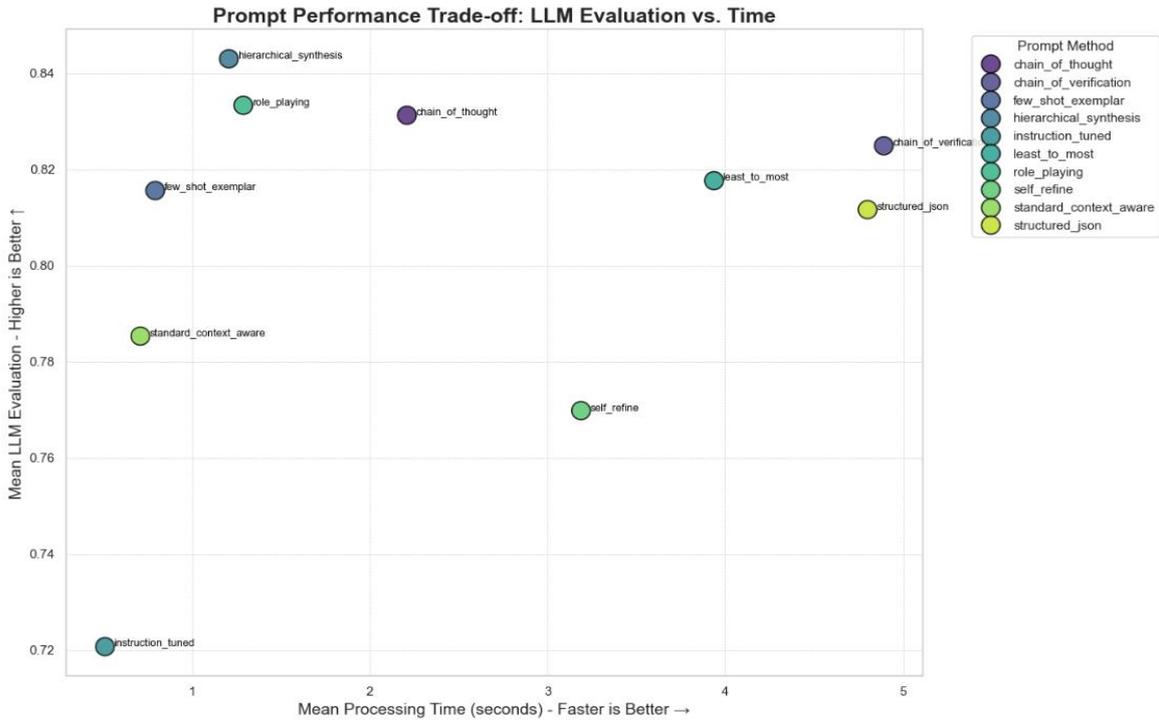

**D.4) Gemma3 4B, baseline + 14 novel prompts:**

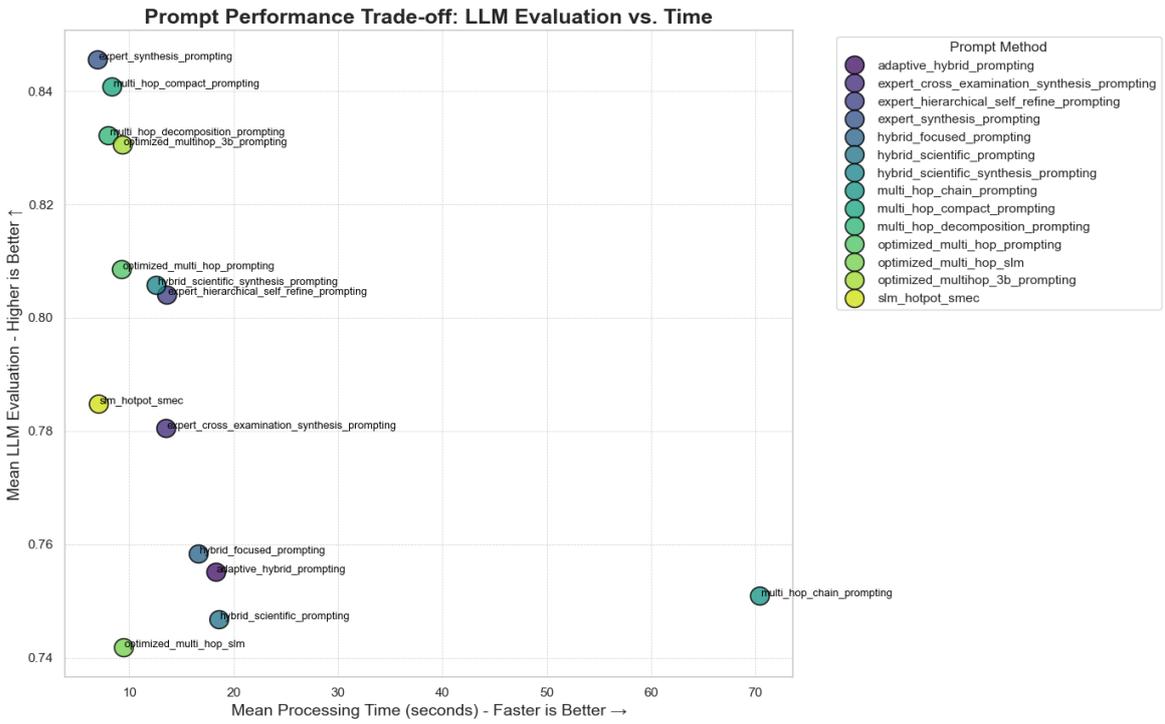

# Appendix E: All Prompts Evaluation

In this section, you can find a complete comparison of all 24 prompts in terms of accuracy, speed, and efficiency.

### E.1) For Qwen2.5 3B:

| Rank | Prompt Method | Accuracy | Time (s) | Efficiency |
|:---:|:---:|:---:|:---:|:---:|
| 1 | optimized_multi_hop_prompting | 0.830 | 13.485 | 0.062 |
| 2 | multi_hop_compact_prompting | 0.825 | 10.145 | 0.081 |
| 3 | multi_hop_decomposition_prompting | 0.806 | 9.952 | 0.081 |
| 4 | multi_hop_chain_prompting | 0.801 | 13.667 | 0.059 |
| 5 | hybrid_scientific_prompting | 0.799 | 17.604 | 0.045 |
| 6 | adaptive_hybrid_prompting | 0.792 | 18.850 | 0.042 |
| 7 | hierarchical_synthesis | 0.788 | 1.621 | 0.486 |
| 8 | optimized_multihop_3b_prompting | 0.786 | 15.144 | 0.052 |
| 9 | hybrid_scientific_synthesis_prompting | 0.783 | 16.303 | 0.048 |
| 10 | structured_json | 0.777 | 1.385 | 0.561 |
| 11 | hybrid_focused_prompting | 0.776 | 14.848 | 0.052 |
| 12 | expert_synthesis_prompting | 0.776 | 12.734 | 0.061 |
| 13 | instruction_tuned | 0.774 | 0.832 | 0.931 |
| 14 | chain_of_thought | 0.773 | 3.090 | 0.250 |
| 15 | chain_of_verification | 0.771 | 3.597 | 0.214 |
| 16 | standard_context_aware | 0.768 | 1.186 | 0.648 |
| 17 | few_shot_exemplar | 0.766 | 1.037 | 0.739 |
| 18 | expert_hierarchical_self_refine_prompting | 0.762 | 14.119 | 0.054 |
| 19 | self_refine | 0.760 | 2.271 | 0.335 |
| 20 | least_to_most | 0.758 | 3.590 | 0.211 |
| 21 | slm_hotpot_smec | 0.758 | 9.851 | 0.077 |
| 22 | role_playing | 0.754 | 1.449 | 0.520 |
| 23 | expert_cross_examination_synthesis_prompting | 0.691 | 19.577 | 0.035 |
| 24 | optimized_multi_hop_slm | 0.682 | 13.795 | 0.049 |

**E.2) For Gemma3 4B:**

| Rank | Prompt Method | Accuracy | Time (s) | Efficiency |
|------|---------------|----------|----------|------------|
| 1 | expert_synthesis_prompting | 0.846 | 6.984 | 0.121 |
| 2 | hierarchical_synthesis | 0.843 | 1.207 | 0.698 |
| 3 | multi_hop_compact_prompting | 0.841 | 8.366 | 0.100 |
| 4 | role_playing | 0.833 | 1.288 | 0.647 |
| 5 | multi_hop_decomposition_prompting | 0.832 | 8.009 | 0.104 |
| 6 | chain_of_thought | 0.831 | 2.209 | 0.376 |
| 7 | optimized_multihop_3b_prompting | 0.831 | 9.377 | 0.089 |
| 8 | chain_of_verification | 0.825 | 4.892 | 0.169 |
| 9 | least_to_most | 0.818 | 3.937 | 0.208 |
| 10 | few_shot_exemplar | 0.816 | 0.793 | 1.029 |
| 11 | structured_json | 0.812 | 4.801 | 0.169 |
| 12 | optimized_multi_hop_prompting | 0.809 | 9.287 | 0.087 |
| 13 | hybrid_scientific_synthesis_prompting | 0.806 | 12.609 | 0.064 |
| 14 | expert_hierarchical_self_refine_prompting | 0.804 | 13.625 | 0.059 |
| 15 | standard_context_aware | 0.785 | 0.709 | 1.107 |
| 16 | slm_hotpot_smec | 0.785 | 7.083 | 0.111 |
| 17 | expert_cross_examination_synthesis_prompting | 0.780 | 13.544 | 0.058 |
| 18 | self_refine | 0.770 | 3.188 | 0.241 |
| 19 | hybrid_focused_prompting | 0.758 | 16.654 | 0.046 |
| 20 | adaptive_hybrid_prompting | 0.755 | 18.338 | 0.041 |
| 21 | multi_hop_chain_prompting | 0.751 | 70.441 | 0.011 |
| 22 | hybrid_scientific_prompting | 0.747 | 18.624 | 0.040 |
| 23 | optimized_multi_hop_slm | 0.742 | 9.485 | 0.078 |
| 24 | instruction_tuned | 0.721 | 0.510 | 1.413 |